\documentclass[11pt]{article}

\usepackage{acl}

\usepackage{latexsym}
\usepackage[T1]{fontenc}
\usepackage{tgheros}
\usepackage{tgcursor}

\usepackage[T1]{fontenc}

\usepackage{enumitem}
\usepackage[utf8]{inputenc}
\usepackage{microtype}

\usepackage{hyperref}

\title{Low-Resource, High-Impact:\\Building Corpora for Inclusive Language Technologies}

\author{
  Ekaterina Artemova \\
  Toloka AI \\
  \small{\texttt{katya-art@toloka.ai}}
  \And
  Laurie Burchell \\
  Common Crawl Foundation \\
  \small{\texttt{laurie@commoncrawl.org}}
  \And
  Daryna Dementieva \\
  TUM \\
  \small{\texttt{daryna.dementieva@tum.de}}
  \AND
  Shu Okabe \\
  TUM \\ 
  \small{\texttt{shu.okabe@tum.de}}
  \And
  Mariya Shmatova \\
  Toloka AI \\
  \small{\texttt{mariya@toloka.ai}}
  \And
  Pedro Ortiz Suarez \\
  Common Crawl Foundation \\
  \small{\texttt{pedro@commoncrawl.org}}
}

\begin{document}
\maketitle
\section{Abstract}
This tutorial\footnote{\href{https://tum-nlp.github.io/low-resource-tutorial}{https://tum-nlp.github.io/low-resource-tutorial}} is designed for NLP practitioners, researchers, and developers working with multilingual and low-resource languages who seek to create more equitable and socially impactful language technologies. Participants will walk away with a practical toolkit for building end-to-end NLP pipelines for underrepresented languages---from data collection and web crawling to parallel sentence mining, machine translation, and downstream applications such as text classification and multimodal reasoning. The tutorial presents strategies for tackling the challenges of data scarcity and cultural variance, offering hands-on methods and modeling frameworks.  
We will focus on fair, reproducible, and community-informed development approaches, grounded in real-world scenarios. We will showcase a diverse set of use cases covering over 10 languages from different language families and geopolitical contexts, including both digitally resource-rich and severely underrepresented languages.

\section{Introduction}

The tutorial is structured into three main parts, each running for about 45 minutes to an hour, followed by a 15-minute Q\&A.
We begin with an introduction to data annotation, where we explore the core tasks, commonly used tools, and typical workflows in NLP (\S\ref{subsec:part1}, Part 1). 
Next, we turn to case studies. Here, we walk through practical pipelines that demonstrate how language technologies are developed for specific languages. The examples span from collecting and crawling language-specific corpora, to building machine translation systems, and finally to transferring knowledge into various downstream applications (\S\ref{subsec:part2}, Part 2).
The final part brings in the voices of experts. Drawing on interviews with 10–15 NLP researchers and practitioners, we share insights into the creation of benchmarks for low-resource and socially impactful settings. This section emphasizes the practical hurdles, design choices, and often overlooked behind-the-scenes aspects of data collection work. (\S\ref{subsec:part3}, Part 3). 

\section{Tutorial Organisers}
\noindent \href{https://scholar.google.ru/citations?user=G0lCb3wAAAAJ}{\textbf{Ekaterina (Katya) Artemova}} is a Professor at the German UDS and a Senior Research Scientist at Toloka AI. Her research focuses on data-centric NLP, with particular emphasis on benchmarking strategies, low-resource settings, and the evaluation of large language models (LLMs).

\noindent \href{https://scholar.google.com/citations?authuser=1&user=11eO-gEAAAAJ}{\textbf{Laurie Burchell}} is a Senior Research Engineer at the Common Crawl Foundation. Their work centres on using data-driven approaches to make language technologies as multilingual as possible, with a focus on high-coverage language identification for large-scale corpus building.

\noindent \href{https://scholar.google.com/citations?user=mLX6olgAAAAJ&hl=en}{\textbf{Daryna Dementieva}} is a postdoctoral researcher at the Technical University of Munich (TUM). Her research focuses on developing robust and linguistically inclusive methods for diverse NLP tasks, with a particular emphasis on advancing resources and technologies for Ukrainian.

\noindent \href{https://aclanthology.org/people/shu-okabe/}{\textbf{Shu Okabe}} is a postdoctoral researcher at TUM Heilbronn. His research interests include developing tools and models for low- and very low-resource languages to support language communities and field linguists.

\noindent \href{https://scholar.google.ru/citations?user=yQ5_EHYAAAAJ&hl=de&oi=ao}{\textbf{Mariya Shmatova}} is the Director of GenAI Data at Toloka AI. Her research focuses on creating datasets for training and evaluating generative AI systems across multiple languages and modalities.

\noindent \href{https://scholar.google.com/citations?user=11eO-gEAAAAJ}{\textbf{Pedro Ortiz Suarez}} is a Principal Research Scientist at the Common Crawl Foundation. His work investigates how data quality affects machine learning model performance and explores data-driven methods to improve these models.

\section{Target Audience}
This tutorial is designed to be introductory and broadly accessible.   Participants are expected to have basic familiarity with data collection and annotation frameworks and a general understanding of core NLP tasks such as text classification, named entity recognition (NER), and machine translation.

The tutorial is aimed at NLP practitioners at various career stages, including students, early-career researchers, and experienced professionals. Participants from diverse backgrounds, ranging from academia and industry to non-profit organizations and low-resource language initiatives, will benefit from the discussion of practical strategies for data collection and model development in low-resource settings.

\section{Outline}

\subsection*{Part 1. Data Annotation Basics}  \label{subsec:part1}

In this section, we introduce key principles, workflows, and best practices in data annotation for NLP, with a focus on quality, scalability, and ethical considerations. We will explore which of these approaches work in a low-resource setting with limited LLM support, and how to overcome potential limitations.

\noindent \textbf{Introduction:} common annotation tasks; annotation lifecycle. 

\noindent \textbf{Designing an annotation task:} common annotation tools; best practices in developing annotation guidelines.

\noindent \textbf{Managing annotators and quality:} recruitment, training and on-boarding of annotators; quality control;  inter-annotator agreement.

\noindent \textbf{Hybrid approaches}: synthetic data; human-in-the-loop annotation with LLMs. 

\subsection*{Part 2. Case studies}  \label{subsec:part2}


\subsubsection*{2.1 Pre-training Data Crawling and Filtering}
\paragraph{Community Annotation for Language Identification of Web Data (paper upcoming)}

This case study describes our experience in creating a corpus to evaluate language identification systems on web data. To do so, we designed a custom annotation tool, focusing on ease of use. We then collaborated with speakers of under-served languages through existing community efforts and organised joint hackathons to create engagement. Finally, we used these annotations to develop a new open dataset for language identification evaluation. This case study addresses the following challenges:
\begin{itemize}[noitemsep,topsep=0pt]
    \item Finding and recruiting suitable annotators
    \item Coordinating the project across multiple academic and industry partners
    \item Designing the task and the annotation interface
    \item Ensuring participant engagement and attribution
    \item Normalising the data to release the final dataset
\end{itemize}

\subsubsection*{2.2 Obtaining Machine Translation System}

\paragraph{Parallel Sentence Mining for Low-Resource Languages \cite{Okabe25improving}}
This case study describes the methodology and challenges when creating a parallel corpus, a valuable resource for downstream NLP tasks such as Machine Translation for low-resource languages. 
To do so, the pairs of sentences are found by mining two monolingual corpora.
We will also detail our work on Upper and Lower Sorbian (\texttt{hsb} and \texttt{dsb}), two low-resource and endangered languages spoken in Germany (Okabe and Fraser, 2025).
\begin{itemize}[noitemsep,topsep=0pt]
    \item Definition, motivation, and challenges of the task of parallel sentence mining, especially for low-resource languages
    \item Creation of a synthetic corpus to evaluate how the current mining pipeline performs
    \item Analysis of the issues that arise, and how they can affect the downstream Machine Translation model
    \item Collaboration with the community: a case study with the Sorbian community.
\end{itemize}

\paragraph{Collecting Low-resource Machine Translation Data}

This case study describes our experience in (1) collecting test sets and (2) evaluating system submissions for low-resource languages for the Machine Translation shared task at WMT 2025. We translated texts from four different domains from English into Bhojpuri (\texttt{bho}), Kenyan Maasai (\texttt{mas}), Serbian (\texttt{sr}), and Egyptian Arabic (\texttt{arz}), and then evaluated these language pairs using the ESA framework \cite{kocmi-etal-2024-error}. 
The case study addresses the following challenges:
\begin{itemize}[noitemsep,topsep=0pt]
\item Recruiting and qualifying for evaluation
\item Handling dialect and script variation
\item Ensuring translation quality, including idiomatic and domain-specific translations
\item Designing and implementing quality control processes
\end{itemize}

\subsubsection*{2.3 Downstream Tasks System Acquisition}

\paragraph{Cross-lingual Text Classification Knowledge Transfer for Low-resource Languages} If an automatic machine translation system is available in a certain way, this case study provides an example of how to obtain a text classification system and data for under-represented languages for various traditional NLP classification tasks. We will cover the case of the Ukrainian language (\texttt{ukr}), obtaining text classification systems for such tasks as toxicity classification, formality classification, and natural language inference (NLI)~\cite{dementieva-etal-2025-cross}. These tasks can play further crucial role in various positive social impact applications such as fake news detection and hate speech mitigation. Thus, this case study will answer the following research questions:
\begin{itemize}[noitemsep,topsep=0pt]
\item Which cross-lingual transfer methods are more efficient? Comparing input machine translation, translation of training data, LLMs prompting, and Adapter Training.
\item How to design semi-synthetic training data, and can it be good?
\item How to design human annotation of the data, and how much data is enough?
\item Finally, what is better: any cross-lingual transfer methods vs. even a small amount of real cultural- and language-specific human-annotated data?
\end{itemize}

\paragraph{Data Labeling for Image Captioning and Visual Question Answering in Low-Resource Dialects \cite{kadaoui2025jeem}}

This case study examines the JEEM benchmark, a culturally grounded and dialect-sensitive dataset designed to evaluate the performance of vision-language models (VLMs) in Arabic-speaking contexts. JEEM focuses on the challenges posed by linguistic and cultural variation across four low-resource Arabic dialects: Jordanian (\texttt{ar-JO}), Egyptian (\texttt{arz}), Emirati (\texttt{ar-AE}), and Moroccan (\texttt{ar-MA}). The dataset supports two core tasks, image captioning and visual question answering (VQA), and focuses on regionally authentic content and dialect-specific annotations. The case study explores key aspects of JEEM's construction, including:
\begin{itemize}[noitemsep,topsep=0pt]
    \item Annotator recruitment and qualification
    \item Curated image sourcing 
    \item Annotation in Modern Standard Arabic (MSA) and dialect
    \item Quality control processes, including manual review and iterative feedback
    \item Human and automatic evaluations to benchmark model performance across metrics such as relevance, fluency, consistency, and dialect authenticity.
\end{itemize}

\subsection*{Part 3. Expert Interviews on Benchmark Creation in Low-Resource Settings} \label{subsec:part3}

We provide an overview of insights gained from interviews with NLP experts actively involved in the creation of benchmarks for positive social impact applications. 

These interviews cover several aspects of using language technologies for social applications, ranging from support for underrepresented languages and dialects to use cases in safety and health. The goal is to explore the often undocumented or underreported aspects of benchmark development, what happens "behind the scenes", including practical challenges, trade-offs between demand and resources, community involvement, and emerging best practices.

We are committed to conducting interviews with 10--15 experts across academia and industry to include a diverse and representative set of perspectives. These conversations aim to complement published literature by shedding light on the real-life experience in collecting data for socially important applications.

\section{Diversity Considerations}

The tutorial centers on the broader goal of achieving true diversity in NLP, with a particular emphasis on collecting data for inclusive language technologies.
We highlight approaches to support the ethical and fair use of human labor in data annotation, aiming both to improve working conditions for annotators and to maintain high data quality.
This tutorial is especially relevant to researchers and activists engaged in data collection for low-resource settings, whether in underrepresented languages or specialized domains.
As outlined in \S\ref{subsec:part3}, Part 3, the included interviews give voice to researchers from a wide range of cultural and geographical backgrounds.
The presenters themselves reflect diversity across gender, language, career stage, background, occupation, affiliation, and location.

\section{Reading List}

Before attending: 
\begin{itemize}[noitemsep,topsep=0pt]
    \item \href{https://toloka.ai/events/toloka-ai-coling-2025-human-w-llm-tutorial}{Tutorial @ COLING'25 on Labelling with LLM and Human-in-the-Loop by Toloka team}
    \item \href{https://sites.google.com/view/nlp4positiveimpact}{NLP for Positive Impact Workshop (NLP4PI)}: a series of workshops to promote NLP application for social good.
\end{itemize}

\noindent Further information: 
\begin{itemize}[noitemsep,topsep=0pt]
    \item \href{https://aclanthology.org/2024.wmt-1.1.pdf}{Findings of the WMT24 General Machine Translation Shared Task: The LLM Era Is Here but MT Is Not Solved Yet}. 
    \item \href{https://www2.statmt.org/wmt25/limited-resources-slavic-llm.html}{LLMs with Limited Resource for Slavic Languages Shared Task @ WMT 2025}: the challenge of machine translation data acquisition for Slavic languages under the condition of the constrained data.
    \item  \href{https://groups.google.com/g/nlp-tools-for-language-communities}{Online Workshop on NLP tools for language communities}: the ongoing project that aims to build a bridge between NLP specialists and language activists.
\end{itemize}

\section{Presenters} 

\href{https://scholar.google.ru/citations?user=G0lCb3wAAAAJ}{\textbf{Ekaterina (Katya) Artemova}} is a Professor at German UDS and a Senior Research Scientist at Toloka AI. She holds a PhD from the FRC CSC RAS. Her research focuses on data-centric NLP, with particular emphasis on benchmarking strategies, low-resource settings, and the evaluation of LLMs. She has co-organized the 1st NLP Power! workshop at ACL'22, a tutorial on artificial text detection at INLG'22, a tutorial on labeling with LLMs and human-in-the-loop at COLING'25, and multiple shared tasks at SemEval'26 and CLEF'25.

\href{https://scholar.google.com/citations?authuser=1&user=11eO-gEAAAAJ}{\textbf{Laurie Burchell}} is a Senior Research Engineer at the Common Crawl Foundation. They hold a PhD in Natural Language Processing from the University of Edinburgh. Their research centres around using data-driven approaches to make language technologies as multilingual as possible, with a particular focus on high-coverage language identification for corpus building. Laurie is a co-organiser for the first Workshop on Multilingual Data Quality Signals at COLM 2025 and a co-organiser of the Open Language Data Initiative which has run two shared tasks as part of WMT 2024 and WMT 2025.

\href{https://scholar.google.com/citations?user=mLX6olgAAAAJ&hl=en}{\textbf{Daryna Dementieva}} is a post-doctoral researcher at Technical University of Munich (TUM). Her PhD topic was ``Methods for Fighting Harmful Multilingual Textual Content'' and she continues her research in responsible AI and NLP. Her current research focuses on developing more robust and linguistically inclusive methods for a range of NLP tasks, with particular emphasis on advancing resources and technologies for Ukrainian. She organized various shared tasks: TextDetox RUSSE-2022, TextDetox CLEF-2024 and CLEF-2025, and recent LLMs with Limited Resources for Slavic Languages WMT-2025. Also, she is one of the core organizers of NLP for Positive Impact Workshop hosted at EMNLP 2024 and ACL 2025 fostering interdisciplinary collaboration with NGOs.

\section{Other Information}

\noindent \textbf{Tutorial type and duration} This is a cutting-edge, half-day, in-person tutorial. 

\noindent \textbf{Approximate count} We expect around 100 attendees, reflecting growing interest in low-resource research at LREC.

\noindent \textbf{Tutorial materials} will be made available online for further use.

\noindent \textbf{Technical Requirements}  We will need a standard room with a projector. 

\section{Ethics Statement}

During the tutorial, we will address key limitations of LLMs for data annotation, including social biases, high compute costs, and limited effectiveness in low-resource settings. We will also stress the responsible employment of human annotators, with fair compensation, informed consent, and reasonable working conditions.

\bibliography{custom}
\end{document}